\documentclass{article}

\usepackage{microtype}
\usepackage{graphicx}
\usepackage{float}
\usepackage{subcaption}
\usepackage{booktabs}
\usepackage{hyperref}

\usepackage[accepted]{icml2026}

\usepackage{amsmath}
\usepackage{amssymb}
\usepackage{xcolor}
\usepackage{tikz}
\usetikzlibrary{positioning, arrows.meta, shapes.geometric, fit, backgrounds}

\icmltitlerunning{Clinical Reasoning Graphs: Competence Without Consistency}

\begin{document}

\twocolumn[
  \icmltitle{Clinical Reasoning Graphs: Structured Evaluation of LLM \\
    Diagnostic Reasoning Reveals Competence Without Consistency}

  \begin{icmlauthorlist}
    \icmlauthor{Nisarg Patel}{ucsf,doc}
  \end{icmlauthorlist}

  \icmlaffiliation{ucsf}{Department of Oral and Maxillofacial Surgery, University of California, San Francisco, CA, USA}
    \icmlaffiliation{doc}{Department of Medicine, Division of Clinical Informatics and Digital Transformation, University of California, San Francisco, CA, USA}
  \icmlcorrespondingauthor{Nisarg Patel}{nisarg.patel@ucsf.edu}

  \icmlkeywords{structured data, clinical reasoning, LLM evaluation, graph representation, healthcare AI}

  \vskip 0.3in
]

\printAffiliationsAndNotice{}

\begin{abstract}
Modern large language models (LLMs) achieve 60-70\% diagnostic accuracy on complex clinical case benchmarks, but accuracy alone cannot distinguish stable clinically-grounded reasoning from pattern matching. We introduce \emph{clinical reasoning graphs}, structured graph representations extracted from free-text LLM diagnostic traces using a domain-grounded ontology with 5 node types and 7 edge types. Applying this extraction pipeline to 750 traces from five LLMs across 50 New England Journal of Medicine Clinicopathological Conference cases and three prompt conditions, we test whether LLM diagnostic traces show stable structured reasoning patterns, i.e., ``diagnostic schemas'', for clinically similar cases, determined by higher graph similarity among clinically similar cases than among clinically dissimilar cases. Across 15 model–condition comparisons, within-cluster and between-cluster composite similarity are nearly equal, with no comparison surviving multiple-testing correction; a component-level analysis finds any residual content signal to be far below schema scale. Graph similarity was also similar for pairs of models that were both correct versus both incorrect, suggesting that graph structure captures a dimension not reflected in diagnosis accuracy. Structured reflection prompting increased explicit discriminating-feature analysis within traces, but did not increase cross-case consistency. We release the ontology, extraction pipeline, validation protocol, and the extracted reasoning graphs and similarity artifacts as resources for structured evaluation of LLM clinical reasoning.
\end{abstract}

\section{Introduction}
\label{sec:intro}

Recent large language models perform increasingly well on complex diagnostic benchmarks
~\citep{eriksen2023nejm, arora2025healthbench}, motivating closer evaluation of not only their final answers but also their reasoning traces. Yet, clinical reasoning traces, the free-text explanations that models produce to support their clinical decisions, remain unstructured, difficult to audit and compare across cases, and difficult to evaluate with traditional accuracy metrics alone.

Clinical reasoning science offers a richer framework for evaluating traces. Expert diagnosticians are thought to organize knowledge through \emph{diagnostic schemas}, which are reusable mental representations that link clinical presentations, enabling conditions, key findings, and discriminating features~\citep{charlin2000scripts, bowen2006educational}. These structures help physicians recognize familiar patterns while updating their reasoning when case-specific findings do not fit a particular framework; for example, a cardiologist encountering chest pain with ST elevation mentally activates a structured pattern of symptom (feature) recognition, hypothesis generation (e.g., myocardial infarction), and discriminating-feature analysis (e.g., which ECG leads show ST elevation?) that is broadly similar for each case.

Here we ask: \emph{do LLM diagnostic traces exhibit anything analogous to diagnostic schemas?} For example, if a model reasons about two nephrotic syndrome cases, do the resulting traces identify overlapping clinical features, differentials, semantic qualifiers, and discriminating features? Or do they follow unrelated reasoning paths that happen to converge on similar final answers?

Answering this requires transforming free-text reasoning traces into structured, computable representations. We introduce \textbf{clinical reasoning graphs}: directed graphs extracted from LLM diagnostic traces using a domain-grounded ontology that captures the entities and relationships of clinical reasoning. Our contributions:
\begin{enumerate}
  \item A \textbf{clinical reasoning ontology} with 5 node types and 7 edge types, grounded in established diagnostic reasoning models, designed to represent features, diagnoses, semantic qualifiers, discriminating features, evidence references, and reasoning-phase transitions
  \item A \textbf{scalable extraction pipeline} converting unstructured traces into validated graphs (750/750 successful, mean 47.0 nodes, 58.7 edges per graph).
  \item An \textbf{empirical test of reasoning consistency}: no evidence of schema-scale cross-case consistency in reasoning traces for clinically similar cases, despite strong diagnostic performance.
  \item Evidence that \textbf{accuracy and reasoning structure capture different evaluation dimensions}: graph similarity is 0.488 when both models are correct versus 0.484 when both are incorrect.
\end{enumerate}

\section{Methods}
\label{sec:methods}

\subsection{Study Design}

We analyze 750 diagnostic reasoning traces generated under a preregistered protocol for diagnostic accuracy and confidence calibration~\citep{patel2026prereg} (generation details in Appendix~\ref{app:generation}). The traces span 50 New England Journal of Medicine (NEJM) Clinicopathological Conference (CPC) cases stratified by physician-assigned difficulty into 15 easy, 20 moderate, and 15 hard cases, and grouped into 20 clinical clusters for the consistency analysis. Five contemporary LLMs were evaluated: GPT-5.4, GPT-5.2, Claude Opus 4.5, Claude Sonnet 4.5, and Gemini 3 Pro. Each case was presented under three prompting conditions. The baseline prompt requested a differential diagnosis and probability estimates directly. The adversarial prompt used a phased protocol and asked the model to argue against its leading diagnosis before offering a revised answer. The structured reflection prompt used a phased protocol requiring a problem representation, balanced stress test, and explicit defend-or-update decision for each diagnosis~\citep{dhaliwal2017art}. Each trace was generated from a single prompt in one autoregressive pass at temperature $1.0$; the phased conditions sequence their protocol phases within that prompt rather than across separate model turns. Final diagnoses were scored against the CPC reference diagnosis by an LLM judge (Claude Opus 4.6).

\subsection{Clinical Reasoning Ontology}
\label{sec:ontology}

We define a directed graph schema grounded in clinical reasoning theory~\citep{charlin2000scripts, bowen2006educational, dhaliwal2017art}:

The ontology contains five node types: (1) clinical features, (2) diagnoses, (3) semantic qualifiers, (4) discriminating features, and (5) evidence references. Clinical features represent signs, symptoms, laboratory values, and imaging findings; diagnoses represent candidate conditions with probability estimates; semantic qualifiers represent abstracted descriptors such as “acute-onset”; discriminating features represent findings used to distinguish between diagnoses; and evidence references represent cited clinical knowledge, such as guidelines, clinical rules, and studies.

The seven edge types capture (1) evidential support, (2) evidence against a diagnosis, (3) discrimination between diagnoses, (4) triggered reflection, and post-reflection diagnosis updates: (5) promoted, (6) demoted, or (7) unchanged.

\begin{figure}[!t]
\centering
\includegraphics[width=\columnwidth]{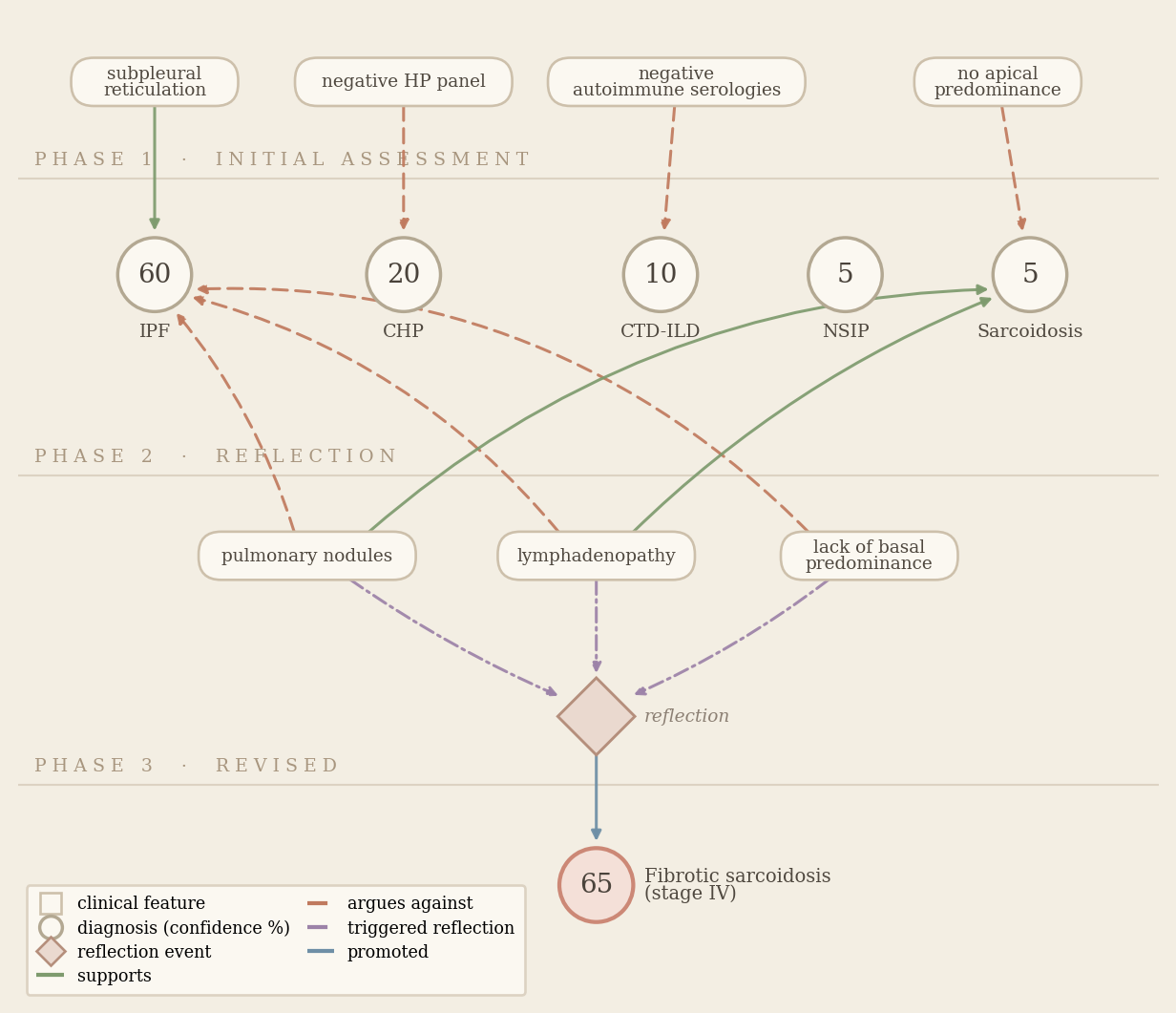}
\caption{A clinical reasoning graph extracted from a single LLM trace,
  read top-down through phases. Features (boxes) support or argue against
  diagnoses (circles; numbers are model confidence). Idiopathic pulmonary
  fibrosis (IPF) leads in Phase~1 ($60\%$); in Phase~2, discriminating
  features (nodules, lymphadenopathy, lack of basal predominance) argue
  against IPF, support sarcoidosis, and trigger a reflection that promotes
  fibrotic sarcoidosis to the revised diagnosis ($65\%$). Abbreviations: HP, hypersensitivity pneumonitis;
  IPF, idiopathic pulmonary fibrosis; CHP, chronic hypersensitivity
  pneumonitis; CTD-ILD, connective tissue disease--associated ILD; NSIP,
  nonspecific interstitial pneumonia.}
\label{fig:graph_example}
\end{figure}

\subsection{Graph Extraction Pipeline}

Each trace was processed by GPT-5.4 using a structured extraction prompt. The prompt specified the ontology, required schema-valid JSON output, enforced connected graphs, required phase tags for all edges, and constrained reflection-related edges in multi-phase traces. The full constraint list appears in Appendix~\ref{app:prompt}. 

All 750 traces yielded valid graphs (100\% extraction rate). Mean statistics: $47.0 \pm 12.4$ nodes, $58.7 \pm 15.2$ edges, mean edge-to-node ratio $1.26$, discriminating edges in 76\% of graphs. Because GPT-5.4 was both an evaluated model and the extraction model, we performed additional checks for source-model bias. Extraction summary statistics did not differ across source models and a subset of traces was independently re-extracted using Claude Opus 4.7.

To assess extraction fidelity, the author, a physician, reviewed all 30 sampled graphs against their source traces. In this sample, edge precision was 98.4\% (1{,}754/1{,}783 edges correct; range 95--100\%). Clinically salient recall, assessed by identifying all explicit reasoning relationships in each source text, was 94.8\% (range 78-97\%; baseline 92.8\%, adversarial 95.5\%, structured 96.2\%). The author did not identify hallucinated edges or missed discriminating edges in the sampled graphs, and phase transitions were correctly captured in the 20 sample traces with multiple phases. The gold (correct) diagnosis was provided to disambiguate edge direction; to test for bias, we re-extracted 20 traces without it. Similarity between gold-included and gold-excluded extractions was $0.862$ (SD~$= 0.060$), exceeding test-retest reliability ($0.825$), with the final diagnosis unchanged in all 20 traces. As an extractor-robustness check, 20 traces were independently re-extracted with Claude Opus 4.7; mean inter-extractor similarity was $0.593$, exceeding within-cluster similarity ($0.456$) and confirming that the extractor does not artificially flatten reasoning differences (Appendix~\ref{app:calibration}).

\subsection{Case Clustering and Similarity Metrics}

We operationalize diagnostic-schema-like consistency as follows: graphs from clinically similar cases should be more similar to one another than graphs from clinically dissimilar cases. To define clinically similar cases, we used a two-stage clustering procedure: organ-system classification followed by embedding-based subclustering (text-embedding-3-large, agglomerative), yielding 20 clusters (44/50 cases in clusters of size $\geq 2$). The author reviewed all clusters for clinical coherence. Within-cluster comparisons are restricted to distinct cases; same-case pairs (the same case reasoned by different models) are excluded, as they measure inter-model agreement rather than cross-case schema reuse. 

Pairwise graph similarity uses a 5-component composite: (1)~\textbf{Feature overlap}: Jaccard on clinical feature labels; (2)~\textbf{Diagnosis overlap}: Jaccard on diagnosis labels; (3)~\textbf{Motif similarity}: $1 - \text{JSD}(p, q)$ on edge-type triplet distributions (source node type $\to$ edge type $\to$ target node type), capturing reasoning style; (4)~\textbf{Qualifier overlap}: Jaccard on semantic qualifiers; (5)~\textbf{Reasoning depth}: the discriminating-edge ratio, defined as the number of discriminating edges divided by the graph's total edge count, with pairwise similarity computed as $1 - |r_a - r_b|$, the absolute difference of the two graphs' ratios. Edges-per-node and reasoning-phase entropy are not part of this component. The primary composite is a weighted mean of these five components (motif similarity 0.35, diagnosis overlap 0.30, feature overlap 0.15, qualifier overlap 0.10, reasoning depth 0.10). These weights follow the original study rather than being re-optimized here; Section~\ref{sec:content_signal} decomposes the composite by component to examine whether the weighting masks a content-specific signal. We also report component-wise results to ensure that the composite does not obscure content-specific effects.
\paragraph{Similarity metric calibration.} Test-retest reliability (same trace extracted twice) yields mean similarity $0.825$ (SD~$= 0.073$, $n = 10$). A noise floor (randomly permuted node labels) yields $0.466$ (SD~$= 0.058$, $n = 50$). These calibration checks provide reference points for interpreting the main analysis; repeated extraction of the same trace produces high similarity, while randomizing node labels yields substantially lower similarity. (Appendix~\ref{app:calibration}).

\subsection{Statistical Analysis}

Primary: within-cluster vs.\ between-cluster composite similarity per model-condition pair. Because pairwise similarities are not independent, all 15 comparisons use case-level permutation tests (10{,}000 permutations preserving model and condition structure) with bootstrap 95\% CIs and Cohen's $d$ effect sizes. Bonferroni correction across 15 comparisons ($\alpha = 0.0033$).

\section{Results}
\label{sec:results}

\subsection{No Evidence of Schema-Scale Cross-Case Consistency}

Within-cluster and between-cluster graph similarity are statistically indistinguishable across all five models and three conditions (Table~\ref{tab:diagnostic_schema} shows the structured condition; full results in Appendix~\ref{app:full_results}). Overall pooled similarity is $0.475$ (within) vs.\ $0.472$ (between).\footnote{This all-condition pooled value is the mean over all 15 model--condition cells (Appendix~\ref{app:full_results}); the structured-condition per-model means in Table~\ref{tab:diagnostic_schema} pool to $0.460/0.460$.} Case-level permutation tests confirmed the null for all 15 comparisons (all $|\Delta| < 0.014$, all $p > 0.02$, none surviving Bonferroni correction; bootstrap 95\% CIs bounded within [$-0.024$, $+0.028$]). Effect sizes are negligible ($d = -0.16$ to $0.26$).

The observed within-cluster similarity under structured reflection ($0.456$) did not exceed the noise floor, a shuffled-label reference point ($0.466$). This suggests that the composite similarity in cross-case comparisons is driven largely by shared graph form rather than shared clinical content. The test-retest ceiling ($0.825$) confirms the metric can detect high similarity when it exists. The primary per-cell test (Table~\ref{tab:diagnostic_schema}) pools pairs within each model--condition cell (same model, distinct cases); the calibration and component ablation (Figure~\ref{fig:accuracy_blind}; Appendix Tables~\ref{tab:calibration}, \ref{tab:ablation}) instead pool across models within the structured condition. Because these are different pair sets, their negligibly small within--between differences need not share a sign: the per-cell pooling leans marginally positive ($+0.003$) and the cross-model pooling marginally negative ($-0.002$), both null and far below the test--retest scale.
\begin{table}[t]
\caption{Diagnostic schema consistency test under structured reflection. Within-cluster (W) vs.\ between-cluster (B) composite similarity. $p$-values from case-level permutation tests (10{,}000 permutations). No comparison survives Bonferroni correction ($\alpha = 0.0033$).}
\label{tab:diagnostic_schema}
\vskip 0.1in
\centering
\small
\begin{tabular}{@{}lcccr@{}}
\toprule
\textbf{Model} & \textbf{W} & \textbf{B} & \textbf{$d$} & \textbf{$p$} \\
\midrule
GPT-5.4        & 0.452 & 0.448 &    0.06 & 0.30 \\
GPT-5.2        & 0.459 & 0.465 & $-$0.11 & 0.80 \\
Sonnet 4.5& 0.460 & 0.463 & $-$0.05 & 0.63 \\
Opus 4.5& 0.474 & 0.469 &    0.09 & 0.22 \\
Gemini 3 Pro   & 0.457 & 0.453 &    0.06 & 0.29 \\
\midrule
\textbf{Pooled}   & \textbf{0.460} & \textbf{0.460} & \textbf{0.01} & \textbf{---} \\
\bottomrule
\end{tabular}
\vskip -0.1in
\end{table}
\paragraph{Robustness.} Component-wise analyses yielded the same qualitative result. The largest within-cluster advantage of any component was $\Delta = +0.003$ (diagnosis overlap); the structural components showed slight between-cluster advantages ($\Delta = -0.006$ for motif similarity, $-0.001$ for depth). The content-sensitive components were especially sparse: feature overlap 0.002 vs. 0.001, diagnosis overlap $0.004$ vs. $0.001$, and qualifier overlap $0.001$ vs. $0.001$ for within- vs. between-cluster comparisons. Structural components were high for both within- and between-cluster pairs, indicating that they capture shared graph form rather than clinical-category similarity. Embedding-based soft matching for synonym variation increased feature overlap but left the within-cluster advantage negligible, $\Delta = +0.002$. Across difficulty tiers, within--between differences ranged from $-0.003$ to $+0.006$, with no consistent direction (Appendix~\ref{app:calibration}). 

\subsection{Accuracy Does Not Predict Graph Similarity}

For each case, we computed graph similarity between all pairs of models that reached the same diagnostic outcome (both correct or both incorrect). Among 1{,}112 such same-outcome pairs, graph similarity when both are correct ($0.488$, $n = 738$) is nearly identical to when both are incorrect ($0.484$, $n = 374$). The difference was not significant (Mann-Whitney $U = 143{,}797$, $p = 0.25$, 95\% CI for $\Delta$: [$-0.003$, $+0.012$]). Accuracy-based evaluation is blind to reasoning dimensions captured by graph-structural analysis.

\subsection{Does the Composite Mask a Content Signal?}
\label{sec:content_signal}

The composite within-versus-between null could, in principle, be an aggregation artifact; a genuine content signal could be diluted by near-ceiling structural components. We test this directly and find that it is not.

\paragraph{The composite is structure-dominated.} Decomposing the composite by weighted contribution, motif similarity and reasoning depth (both near ceiling, $\approx 0.88$ and $\approx 0.98$, and statistically indistinguishable within versus between) account for $\approx 89\%$ of the composite's value while carrying $45\%$ of its weight. A further $\approx 11\%$ comes from the qualifier component scoring empty-versus-empty comparisons as identical ($587/750$ graphs contain no semantic qualifier). Realized clinical-content overlap (feature and diagnosis) contributes $\approx 0.3\%$ of the composite's level. The aggregate is therefore almost entirely a structural-ceiling-plus-empty-set quantity, content-insensitive by construction.

\paragraph{Isolating content does not reveal a schema-scale signal.} Recomputing similarity on the content channel alone (the mean of feature, diagnosis, and qualifier Jaccard, with empty-versus-empty scored $0$), within-cluster similarity is $0.0021$ versus $0.0006$ between-cluster. The within-cluster contrast is statistically detectable, it is positive in all 15 model--condition cells, and 7 of 15 survive Bonferroni correction under a case-level permutation test. But, the within-cluster level sits at the metric's noise floor (shuffled-label floor $0.0006$--$0.0014$; within exceeds it by only $0.12$--$0.37$ floor SDs) and reaches just $0.5\%$ of the content channel's own test-retest ceiling ($0.43$). A statistically significant contrast between two near-floor quantities is not a signal above the floor; within-cluster content overlap does exceed between-cluster overlap, but both sit essentially at the noise floor, so the excess is real yet marginal, not schema-scale.

\paragraph{A weak, cluster-independent association.} To remove circularity with cluster construction, we regress pairwise graph \emph{content} similarity (the content channel, with empty-versus-empty qualifier pairs scored $0$) on pairwise gold-diagnosis similarity (local PubMedBERT embeddings of the reference diagnoses; a Mantel test per cell, with a case-resampling bootstrap). This isolates clinical-content agreement from the near-ceiling structural and empty-qualifier components that dominate the full composite. The content channel shows a positive but weak association (mean Mantel $r = 0.118$, $\approx 1.4\%$ of variance, positive in all 15 cells; pooled bootstrap slope $+0.0037$, $95\%$ CI $[+0.0020, +0.0057]$, excluding zero). This is the cleanest evidence for any residual content signal: real, but far below schema scale.

\paragraph{Interpretation.} These analyses reinforce rather than overturn the primary result. The composite null is not an aggregation artifact because the content channel it could be masking is itself at the noise floor. We bound the claim to what the instrument measures. The content metric's test-retest reliability is only $0.43$, which attenuates any true correlation, meaning this is evidence against \emph{detectable} stable diagnostic schemas in these graph representations,  but not proof that no latent organizing structure exists. Establishing the latter would require a higher-reliability representation (e.g., embedding-based soft matching) and a consensus human-clinician baseline.

\subsection{Structured Reflection Increases Discriminating Structure, Not Consistency}

The preceding analyses measure reasoning structure \emph{across} traces. We now examine discriminating structure within individual traces.

Structured reflection produces substantially more within-trace discriminating structure (more Phase 2 reasoning edges and discriminating motifs, +33\%, $p < 10^{-6}$), confirming the pipeline detects genuine cross-condition structural variation. However, structured reflection did not increase within-cluster consistency; relative to baseline, per-model within-cluster composite similarity fell for four of five models and rose only negligibly for the fifth (GPT-5.2), with the per-model change spanning $-0.041$ to $+0.005$ and no model showing a significant increase (all $p > 0.58$). Structured reflection introduces case-specific discriminating analysis that \emph{diversifies} rather than standardizes graph structure.

Cross-model consistency is also flat. Under structured reflection, within-cluster similarity spans only $0.452$--$0.474$ across the five models.

\begin{figure}[!t]
\centering
\includegraphics[width=\columnwidth]{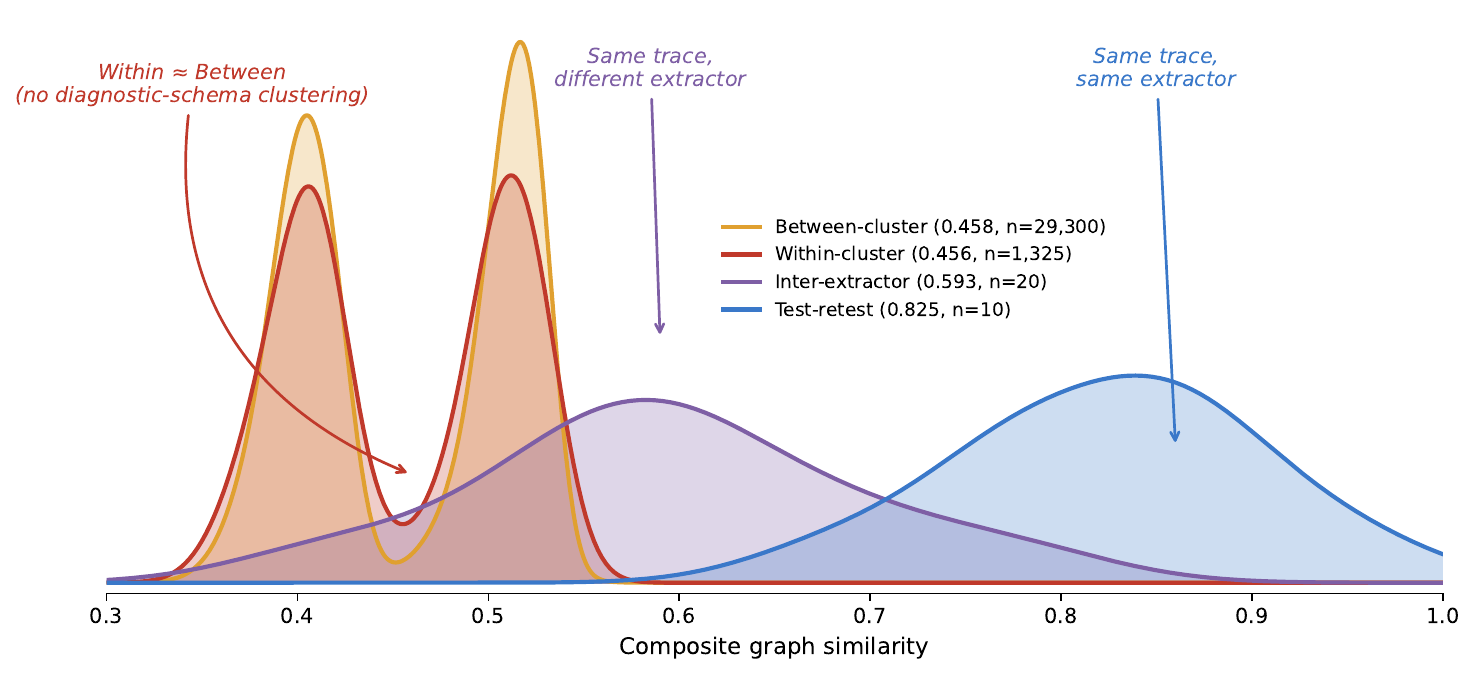}
\caption{Empirical distributions of pairwise graph similarity. Within-cluster and between-cluster distributions overlap almost completely (no detectable diagnostic-schema-like clustering); their shared bimodality reflects the qualifier component (Appendix~\ref{app:calibration}), not clinical structure. Inter-extractor (same trace, different extraction model) and test-retest (same trace, same model) distributions are clearly separated, confirming the metric detects similarity when it should exist.}
\label{fig:accuracy_blind}
\vspace{-0.5em}
\end{figure}

\section{Related Work}
\label{sec:related}

\paragraph{Chain-of-thought evaluation.}
Recent work examines whether CoT reasoning is faithful~\citep{lanham2023measuring, turpin2024language}. \citet{basu2026illusion} show that individual reasoning steps are often decorative, i.e., removable without changing the answer, establishing that reasoning can be unfaithful \emph{within} a trace. Our work reveals a complementary problem, showing that reasoning is inconsistent \emph{across} traces for clinically similar cases.

\paragraph{Clinical LLM benchmarks.}
MedQA~\citep{jin2021medqa}, MedPaLM~\citep{singhal2023large}, and HealthBench~\citep{arora2025healthbench} evaluate accuracy on clinical tasks. Our result suggests that diagnostic accuracy does not, by itself, characterize the graph structure of generated reasoning traces, arguing for complementary process-level evaluation.

\paragraph{Diagnostic schemas in clinical education.}
\citet{mccoy2025sct} recently showed that LLMs perform markedly worse on script concordance testing than on standard benchmarks. Our work complements this by testing whether graph-structural consistency serves as a proxy for stable reasoning representations.

\section{Discussion}
\label{sec:discussion}

\paragraph{Structured adaptation as evaluation infrastructure.}
Our pipeline pattern -- define a domain ontology, extract structured graphs from free-text reasoning, and validate the resulting objects -- could be adapted to other domains. The 0.488 vs.\ 0.484 result ($p = 0.25$) suggests that accuracy and graph structure capture different aspects of model behavior; two models with identical benchmark scores may produce substantially different reasoning structures. For clinical AI evaluation, final-answer accuracy should therefore be complemented by process-level analyses that assess how conclusions are generated and revised. Additionally, as clinical systems grow more agentic, the same representation could extend to multi-step agent trajectories, representing actions, tool calls, and state transitions as comparable graphs (with the ontology extended for action sequences). 

\paragraph{Implications for process-level evaluation.}
LLMs achieve diagnostic competence without the schema-scale cross-case reasoning consistency characteristic of expert reasoning~\citep{charlin2000scripts}. This interpretation is consistent with the hypothesis that current LLM traces resemble case-specific reconstruction more than reusable expert-like reasoning templates, but awaits direct comparison with expert physicians on the same cases. If reasoning traces vary substantially across clinically similar cases, then auditing a model on one representative case may not generalize to nearby cases.

\paragraph{Limitations.}
Clinical reasoning graphs operationalize expressed reasoning, not internal model cognition, and the ontology may miss aspects of diagnostic schema structure that clinicians would recognize. The 50-case corpus limits subgroup power, and we lack a consensus human clinician baseline. The null holds for both pre-2023 ($\Delta < 0.001$) and post-2023 cases ($\Delta = +0.005$), reducing concern about LLM memorization bias.


\section*{Software and Data}
The evaluation code, clinical reasoning graph ontology and extraction pipeline, validation protocol, extracted reasoning graphs, pairwise similarity arrays, gold-diagnosis embeddings, and case-level metadata are available at \url{https://github.com/nisargpatel/clinical-reasoning-graphs}. Reasoning traces are released with verbatim NEJM case text removed; the underlying CPC cases are published in the \emph{New England Journal of Medicine} and are not redistributed here. Two embedding models are used and reported per analysis, the OpenAI text-embedding-3-large (3072-dimensional) for embedding-based subclustering and feature-level soft matching and a local PubMedBERT model (\texttt{NeuML/pubmedbert-base-embeddings}, 768-dimensional) for the gold-diagnosis embeddings in the Mantel and content-channel regressions. The 750 traces were generated under the preregistered protocol~\citep{patel2026prereg}.

\section*{Impact Statement}
As language models increasingly surface model ``reasoning'' -- chain-of-thought, structured rationales -- there is a growing temptation to ground trust in its apparent coherence, treating reasoning that resembles a clinician's diagnostic schema as more trustworthy. This work shows that diagnostic competence and reasoning-structure consistency are dissociable. Models that reach accurate diagnoses do not produce reasoning graphs that are measurably more similar for clinically similar cases than for dissimilar ones, and what cross-case similarity exists is governed largely by graph form and scoring conventions rather than clinical content. The practical implication is that aggregate reasoning-similarity can appear consistent for reasons unrelated to clinical reasoning, so ``the reasoning looks consistent'' is not, on its own, evidence of reliable or reusable clinical reasoning , which is a distinction that matters when such signals inform clinician trust or automated triage. We frame this as a measurement result, evidence against detectable diagnostic-schema consistency in these representations under a metric of limited reliability rather than proof that no organizing structure exists. More constructively, representing free-text reasoning as structured graphs converts an opaque artifact into one that can be inspected, compared, and contested; we release the ontology, extraction pipeline, and artifacts so that claims about reasoning consistency can be tested rather than assumed.

\section*{Acknowledgements}
We thank Gurpreet Dhaliwal for valuable feedback on this work and the anonymous SD4H reviewers for constructive feedback that improved this work.

\bibliography{references}
\bibliographystyle{icml2026}

\appendix
\section{Generation Details}
\label{app:generation}

The 750 reasoning traces analyzed here were produced under the preregistered protocol~\citep{patel2026prereg} on 9-10 March 2026, with all models queried through the OpenRouter API at temperature $1.0$. Each trace is a single autoregressive generation from one prompt; the phased conditions sequence their protocol phases within that prompt. Five models were evaluated as test subjects (Table~\ref{tab:gen_models}). A separate chairman model (\texttt{anthropic/claude-opus-4.6}), used only as an evaluator and never as a test subject, scored each trace's final diagnosis against the published CPC reference: whether the leading diagnosis was correct (top-1) and whether the reference appeared in the top-3 and top-5 differential. The full generation and scoring prompts are included in the released repository.

\begin{table}[h]
\caption{Models evaluated, with the OpenRouter identifiers used for this run (9--10 March 2026). Model versions in any later run of the companion study may differ.}
\label{tab:gen_models}
\centering
\small
\begin{tabular}{@{}ll@{}}
\toprule
\textbf{Model} & \textbf{OpenRouter ID} \\
\midrule
GPT-5.4           & \texttt{openai/gpt-5.4} \\
GPT-5.2           & \texttt{openai/gpt-5.2} \\
Claude Opus 4.5   & \texttt{anthropic/claude-opus-4.5} \\
Claude Sonnet 4.5 & \texttt{anthropic/claude-sonnet-4.5} \\
Gemini 3 Pro      & \texttt{google/gemini-3-pro-preview} \\
\bottomrule
\end{tabular}
\end{table}

\section{Graph Extraction Constraints}
\label{app:prompt}

The extraction prompt enforces the following schema and consistency constraints: (1)~every node has $\geq$1 edge; (2)~multi-phase traces include $\geq$1 \textsc{triggered-reflection} edge; (3)~extracted edges must correspond to explicit reasoning relationships in the trace; (4)~every edge phase-tagged; (5)~discriminating features connect $\geq$2 diagnoses; (6)~probability estimates extracted verbatim; (7)~promoted/demoted edges require probability changes; (8)~semantic qualifiers abstract from raw findings; (9)~evidence references cite specific knowledge; (10)~no duplicate node labels; (11)~the graph must form a single connected component, unless the trace explicitly contains unrelated reasoning fragments.

\section{Clinical Category Clusters}
\label{app:clusters}

The 14 multi-case clusters span 12 organ-system categories, with subclustering where clinically appropriate: neurologic (2 subclusters, $n = 7$), cardiovascular (2 subclusters, $n = 6$), pulmonary ($n = 5$), infectious disease ($n = 4$), hematologic ($n = 4$), dermatologic ($n = 3$), rheumatologic/autoimmune ($n = 3$), ophthalmologic ($n = 3$), hepatic ($n = 3$), endocrine ($n = 2$), gastrointestinal ($n = 2$), and renal ($n = 2$), plus 6 singletons (20 clusters total). 44/50 cases contribute to within-cluster analysis.

\section{Full Diagnostic Schema Consistency Results}
\label{app:full_results}

Table~\ref{tab:full_results} reports within-cluster (W) vs.\ between-cluster (B) composite similarity across all 15 model--condition comparisons. Each comparison has $n_{\text{within}} = 53$ and $n_{\text{between}} = 1{,}172$ pairs.

\begin{table}[H]
\caption{Full diagnostic schema consistency results. $p$-values from case-level permutation tests (10{,}000 permutations). Asterisks indicate nominal $p < 0.05$; no comparison survives Bonferroni correction ($\alpha = 0.0033$).}
\label{tab:full_results}
\vskip 0.1in
\centering
\scriptsize
\begin{tabular}{@{}llcccr@{}}
\toprule
\textbf{Model} & \textbf{Cond.} & \textbf{W} & \textbf{B} & \textbf{$d$} & \textbf{$p$} \\
\midrule
GPT-5.4      & base.      & 0.476 & 0.469 &    0.11 & 0.18 \\
GPT-5.4      & adv.       & 0.475 & 0.483 & $-$0.16 & 0.92 \\
GPT-5.4      & struct.    & 0.452 & 0.448 &    0.06 & 0.30 \\
\addlinespace
GPT-5.2      & base.      & 0.454 & 0.462 & $-$0.14 & 0.88 \\
GPT-5.2      & adv.       & 0.457 & 0.458 & $-$0.03 & 0.57 \\
GPT-5.2      & struct.    & 0.459 & 0.465 & $-$0.11 & 0.80 \\
\addlinespace
Sonnet 4.5   & base.      & 0.501 & 0.498 &    0.05 & 0.34 \\
Sonnet 4.5   & adv.       & 0.500 & 0.492 &    0.17 & 0.09 \\
Sonnet 4.5   & struct.    & 0.460 & 0.463 & $-$0.05 & 0.63 \\
\addlinespace
Opus 4.5     & base.      & 0.495 & 0.481 &    0.22 & 0.038* \\
Opus 4.5     & adv.       & 0.494 & 0.485 &    0.17 & 0.09 \\
Opus 4.5     & struct.    & 0.474 & 0.469 &    0.09 & 0.22 \\
\addlinespace
Gemini 3     & base.      & 0.482 & 0.469 &    0.26 & 0.021* \\
Gemini 3     & adv.       & 0.492 & 0.484 &    0.15 & 0.11 \\
Gemini 3     & struct.    & 0.457 & 0.453 &    0.06 & 0.29 \\
\midrule
\textbf{Pooled} & \textbf{all} & \textbf{0.475} & \textbf{0.472} & \textbf{0.04} & --- \\
\bottomrule
\end{tabular}
\vskip -0.1in
\end{table}

\section{Similarity Metric Calibration and Ablation}
\label{app:calibration}

 \begin{table}[H]
  \caption{Calibration bounds for the similarity metric, for the weighted
  composite and the content-only channel (mean of feature, diagnosis, and
  qualifier Jaccard, empty-versus-empty scored $0$); SD pertains to the
  composite. Conditions: \emph{noise floor}, shuffled node labels;
  \emph{between-/within-cluster}, dissimilar/similar cases;
  \emph{inter-extractor}, GPT-5.4 vs.\ Claude Opus 4.7 re-extraction;
  \emph{test-retest}, same trace re-extracted; \emph{gold-ablation},
  extraction with vs.\ without the gold diagnosis.}
  \label{tab:calibration}
  \vskip 0.1in
  \centering
  \scriptsize
  \begin{tabular}{@{}lcccr@{}}
  \toprule
  \textbf{Condition} & \textbf{Composite} & \textbf{Content} & \textbf{SD} & \textbf{$n$} \\
  \midrule
  Noise floor      & 0.466 & 0.0006--0.0014 & 0.058 & 50 \\
  Between-cluster  & 0.458 & 0.0007 & ---    & 29{,}300 \\
  Within-cluster   & 0.456 & 0.0021 & ---    & 1{,}325 \\
  Inter-extractor  & 0.593 & 0.125  & 0.094  & 20 \\
  Test-retest      & 0.825 & 0.432  & 0.073  & 10 \\
  Gold-ablation    & 0.862 & 0.514  & 0.060  & 20 \\
  \bottomrule
  \end{tabular}
  \vskip 0.1in
  \end{table}

\begin{table}[H]
\caption{Similarity ablation by individual component (cross-model, structured condition). Component rows report realized set-Jaccard overlap (empty-versus-empty qualifier pairs scored $0$) and weight-average to $\approx 0.405$; the composite ($0.456$) is higher because its qualifier component instead scores empty-versus-empty pairs as $1.0$ ($\approx 51\%$ of structured pairs lack a qualifier in both traces), adding $\approx 0.051$. This cross-model pooling also differs from the per-cell pooling in Table~\ref{tab:full_results} ($0.475/0.472$). The within--between difference is negligible for every component.}
\label{tab:ablation}
\vskip 0.1in
\centering
\small
\begin{tabular}{@{}lccc@{}}
\toprule
\textbf{Component} & \textbf{Within} & \textbf{Between} & \textbf{$\Delta$} \\
\midrule
Feature overlap      & 0.002 & 0.001 & $+$0.001 \\
Diagnosis overlap    & 0.004 & 0.001 & $+$0.003 \\
Motif similarity     & 0.875 & 0.881 & $-$0.006 \\
Qualifier overlap    & 0.001 & 0.001 &    0.000 \\
Depth similarity     & 0.976 & 0.977 & $-$0.001 \\
\midrule
Composite            & 0.456 & 0.458 & $-$0.002 \\
Soft Jaccard (emb.)  & 0.003 & 0.001 & $+$0.002 \\
\bottomrule
\end{tabular}
\vskip -0.1in
\end{table}

Content-sensitive components (feature and diagnosis Jaccard) show near-zero overlap in both conditions. Structural components (motif distribution, reasoning depth) were high for both within- and between-cluster comparisons, suggesting that they capture shared prompt and ontology structure rather than case-specific clinical similarity. For this reason, we report content-sensitive components separately.

\paragraph{Bimodality.} The composite distribution is bimodal (Figure~\ref{fig:accuracy_blind}, modes near $0.41$ and $0.51$), which reflects the qualifier component rather than clinical structure. A semantic-qualifier node is present in only $28\%$ of structured graphs (versus $20\%$ baseline, $16\%$ adversarial), so $51\%$ of structured pairs contain no qualifier in either trace; set-Jaccard scores such empty-versus-empty pairs $1.0$, placing them in the upper mode, even though the realized qualifier overlap reported in Table~\ref{tab:ablation} is near zero. Because the component is effectively binary ($\approx 1.0$ when both traces lack a qualifier, $\approx 0$ otherwise), the mode separation equals its $0.10$ weight. The component is nearly symmetric across the contrast (both-empty rate $0.50$ within versus $0.51$ between), so it confers no within-cluster advantage; its slight residual runs toward between-cluster pairs, so it cannot mask a within-cluster signal. The small residual within-versus-between difference is itself partly a qualifier artifact, not evidence of cross-case schema similarity. Importantly, qualifier nodes are not exclusive to structured reflection; they appear, sparsely, under every condition.

\paragraph{Semantic matching.} Recomputing feature overlap with embedding-based soft matching (text-embedding-3-large, cosine threshold $0.85$) increases within-cluster feature overlap from $0.002$ to $0.003$ ($\Delta = +0.002$ vs.\ between-cluster). The absolute overlap remains near zero; models identify largely non-overlapping clinical features even for cases in the same clinical category. The within-cluster advantage ($0.002$) is negligible relative to the test-retest ceiling ($0.825$).

\paragraph{Inter-extractor agreement.} Twenty traces (from the clinician validation sample) were re-extracted using Claude Opus 4.7 and compared to GPT-5.4 extractions. Mean inter-extractor similarity was $0.593$ (SD~$= 0.094$), indicating substantial extractor-dependent variation. Within-cluster similarity ($0.456$) falls well below inter-extractor agreement ($0.593$). Case-to-case variation in LLM reasoning exceeds extractor-to-extractor variation on the same trace. These results reduce the concern that GPT-5.4 extraction alone explains the null result. 

\section{Extraction Prompt}
\label{app:extraction_prompt}

The extraction system prompt instructs GPT-5.4 to extract a JSON reasoning graph from each trace. The prompt defines the ontology (Section~\ref{sec:ontology}) and enforces the 11 constraints (Appendix~\ref{app:prompt}). Key structural elements:

\begin{quote}
\footnotesize
\textbf{Node types:} \texttt{clinical\_feature} (verbatim terms from trace), \texttt{diagnosis} (with integer confidence 0--100), \texttt{semantic\_qualifier} (abstracted framing descriptors), \texttt{discriminating\_feature} (features explicitly identified as distinguishing between diagnoses), \texttt{evidence\_reference} (cited guidelines, studies, clinical rules).

\textbf{Edge types:} \texttt{supports} (feature $\to$ diagnosis), \texttt{argues\_against} (feature $\to$ diagnosis), \texttt{discriminates\_between} (feature $\to$ [dx\_A, dx\_B]), \texttt{triggered\_reflection} (feature $\to$ reflection event), \texttt{promoted}/\texttt{demoted}/\texttt{unchanged} (reflection event $\to$ diagnosis, with confidence\_before and confidence\_after metadata).

\textbf{Critical rules:} (1)~Extract every reasoning relationship, not just major ones. (2)~Preserve verbatim language for each edge. (3)~Do not infer relationships the model did not explicitly state. (4)~Zero orphan nodes -- every node must connect to $\geq$1 edge. (5)~Reflection events mandatory for multi-phase traces. (6)~Edge count should be $\geq$ node count. (7)~Phase-specific header detection accounts for variation across models (e.g., ``Stress Test Your Leading Diagnosis'' vs.\ ``Strongest argument AGAINST'').
\end{quote}

\noindent The user prompt provides the case presentation, correct diagnosis (for reference only), model name, condition, and the full reasoning trace. The full prompt (238 lines) is available in the code repository (\url{https://github.com/nisargpatel/clinical-reasoning-graphs}).

\end{document}